\pdfoutput=1

\documentclass[11pt]{article}

\usepackage[table]{xcolor}
\usepackage{acl}

\usepackage{times}
\usepackage{latexsym}

\usepackage[T1]{fontenc}

\usepackage[utf8]{inputenc}

\usepackage{microtype}

\usepackage{graphicx}

\usepackage{booktabs}
\usepackage{multirow}
\usepackage{todonotes}
\usepackage{amsmath}
\usepackage[normalem]{ulem}
\usepackage{caption}
\usepackage{subcaption}

\newcommand{\mbert}[0]{mBERT}

\newcommand{\numoflangs}[0]{22}
\newcommand{\draftcomment}[3]{{\textcolor{#3}{[#1]#2}}}
\renewcommand{\draftcomment}[3]{}  

\newcommand{\taskeval}[0]{\varepsilon}
\newcommand{\zs}[0]{\mathcal{Z}}
\newcommand{\ft}[0]{\mathcal{F}}
\newcommand{\donation}[0]{\mathcal{D}}
\newcommand{\recipience}[0]{\mathcal{R}}
\newcommand{\clt}[3]{\zs_{#3}(#1\rightarrow #2)}
\newcommand{\ftt}[2]{\ft(#1 \rightarrow #2)}
\newcommand{\xtreme}[0]{XTREME}

\newcommand{\repourl}[0]{github.com/SLAB-NLP/linguistic-blood-bank}
\newcommand{\repo}[0]{\url{\repourl}}
\newcommand{\demo}[0]{\url{\repourl\#interactive-exploration}}

\definecolor{bottlegreen}{rgb}{0.0,0.42,0.31}
\definecolor{donorred}{RGB}{228.,116.,95.}
\definecolor{reciepientblue}{RGB}{0,152,251}

\title{A Balanced Data Approach for Evaluating Cross-Lingual Transfer: \\ Mapping the Linguistic Blood Bank}

\author{Dan Malkin$^{\diamondsuit}$ \qquad
Tomasz Limisiewicz$^{\spadesuit}$\thanks{$\;\;$Work done while visiting the Hebrew University.} \qquad
  Gabriel Stanovsky$^{\diamondsuit}$ \\
  $^{\diamondsuit}\;$School of Computer Science, The Hebrew University of Jerusalem \\
  $^{\spadesuit}\;$Faculty of Mathematics and Physics, Charles University in Prague
 \\
  \texttt{\{dan.malkinhueb,gabriel.stanovsky\}@mail.huji.ac.il} \\
  \texttt{limisiewicz@ufal.mff.cuni.cz}}

\begin{document}
\maketitle
\begin{abstract}
We show that the choice of pretraining languages affects downstream cross-lingual transfer for BERT-based models. We inspect zero-shot performance in balanced data conditions to mitigate data size confounds, classifying pretraining languages that improve downstream performance as \emph{donors}, and languages that are improved in zero-shot performance as \emph{recipients}. We develop a method of quadratic time complexity in the number of languages to estimate these relations, instead of an exponential exhaustive computation of all possible combinations. 
We find that our method is effective on a diverse set of languages spanning different linguistic features and two downstream tasks.
Our findings can inform developers of large-scale multilingual language models in choosing better pretraining configurations.\footnote{Code and models are publicly available at: \repo{}}

\end{abstract}

\section{Introduction}

\begin{figure}[t!]
    \centering
    
    \includegraphics[width=0.45\textwidth]{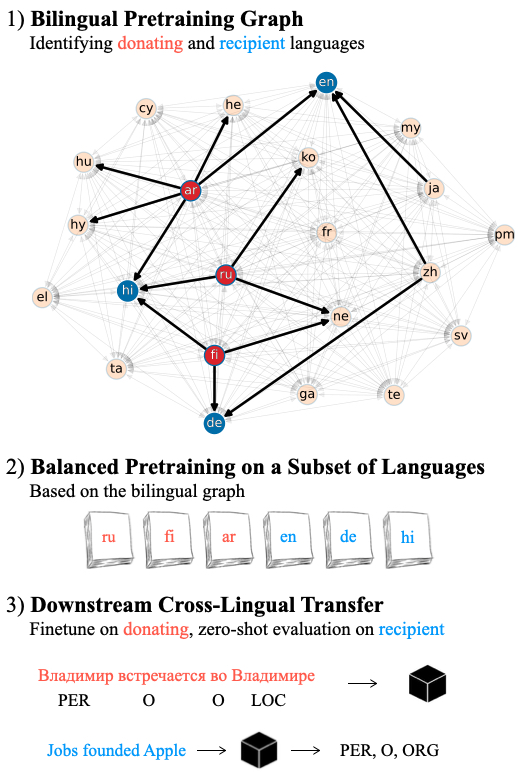}
    \caption{
    We build a complete, directed graph over a diverse set of 22 languages. Weighted edges show the improvement of bilingual LM over monolingual performance (bold edges represent larger weights).
    Languages which consistently improve performance are termed ``donors'' and marked in \textcolor{donorred}{red}, while languages 
    which benefit most are termed ``recipients'' (marked in \textcolor{reciepientblue}{blue}).
    We show that our observations hold in several configurations on two downstream tasks.
    \label{fig:schwartz}}
\end{figure}

Pretrained language models are setting state-of-the-art results by leveraging raw texts during pretraining~(PLMs; \citealp[inter alia]{elmo,devlin-etal-2019-bert}).
Interestingly, when pretraining on multilingual corpora, PLMs seem to exhibit \emph{zero-shot} cross-lingual abilities, achieving non-trivial performance on downstream examples in languages seen only during pretraining. For example, in the bottom of Figure \ref{fig:schwartz}, a named entity recognition model finetuned on Russian is capable of predicting correctly name entity tags for texts in English, seen only during pretraining~\citep{pires-etal-2019-multilingual,wu2019emerging,wang2019cross,conneau2019unsupervised,lazar-etal-2021-filling,turc2021revisiting}.

Previous analyses examined how several factors contribute to this emerging behavior. For example, parameter sharing and model depth are important in certain configurations~\citep{wang2019cross,wu2019emerging},
as well as typological similarities between languages~\citep{pires-etal-2019-multilingual}, and the choice of specific finetune languages~\citep{turc2021revisiting}.



In this work, we focus on an important factor that we find missing in prior work, namely the effect that \emph{pretraining} languages have on downstream zero-shot performance.
In particular, we ask three major research questions: 
(1)~Does the choice of pretraining languages affect downstream cross-lingual transfer, and if so, to what extent?
(2)~Is English the optimal pretraining language, when controlling for confounding factors such as data size and domain? And finally, 
(3)~Can we choose pretraining languages to improve downstream zero-shot performance? 


In addressing these research questions, we aim to decouple  the \emph{language} from its corresponding \emph{dataset}. To the best of our knowledge, prior work has conflated pretrain corpus size and its domain with other examined factors, thus skewing results towards over-represented languages, such as English or German~\citep{joshi-etal-2020-state}.\footnote{For example, English was X100 more likely to be sampled in \mbert{}'s pretraining data than Icelandic.}
To achieve this, we first construct a \emph{linguistically-balanced} pretraining corpus based on Wikipedia, composed of a diverse set of \numoflangs{} languages. We carefully control for the amount of data and domain distribution in each of the languages~(Section~\ref{sec:data}). 

Next, since the number of pretraining configurations grows  exponentially with the number of languages $n$ represented in the dataset, it is infeasible to exhaustively test all possible configurations, much less extend it for more languages.\footnote{There are $2^n$ possible pretraining configurations taking into account inclusion and omission of every language.}
In Section~\ref{sec:language-graph} we propose a novel pretraining-based approach that is quadratic in the number of languages. This is achieved by training all $\binom{n}{2}$ combinations of bilingual masked language models over our corpus, thus yielding a complete directed graph~(Figure~\ref{fig:schwartz}), where an edge $l_1 \to l_2$ estimates how much a language $l_1$ contributes to zero-shot performance in language $l_2$, based only on language modeling performance.

In Section~\ref{sec:pretraining-analysis}, we use the graph to identify languages which generally contribute as pretraining languages (termed ``donors''), and languages which often benefit from 
training with other languages (termed ``recipients''). Further, we use the graph to make  observations regarding the effect of typological features on bilingual language modeling, and make available an interactive graph explorer.


Finally, our evaluations on two multilingual downstream tasks (part of speech tagging and named entity recognition) lead to three main conclusions~(Section~\ref{sec:downstream}): 
(1) the choice of pretraining languages indeed leads to differences in zero-shot performance; 
(2) controlling for the amount of data allotted for each language during pretraining questions the primacy of English as the main \emph{pretraining} language; and 
(3) our hypotheses regarding donors and recipient language hold in both downstream tasks, and against two additional control groups. 

\section{Metrics for Pretraining-Aware Cross-Lingual Transfer}
\label{sec:definitions}
In this section, we extend existing metrics for zero-shot cross-lingual transfer to account for \emph{pretraining} languages. 
Intuitively, our metrics for a model $M$ and a given downstream task take into account three factors: (1) $P$, the set of languages seen during pretraining, (2) $s \in P$, the \emph{source} language used for finetuning, and (3) $t \in P$, the \emph{target} language, seen during inference.

Formally, we adapt the formulation of \citet{hu2020xtreme} to define a \emph{pretraining-aware} bilingual zero-shot transfer score $\mathcal{Z}$ as:\footnote{We opt not to normalize the score by the monolingual performance as done in~\citet{turc2021revisiting}, as we do not want it to affect the score.}

\begin{equation}
\label{eq:taskeval}
    \clt{s}{t}{P} := \taskeval{}(M^{P,s}, t) 
\end{equation}

Where $M^{P,l}$ is a model pretrained on the set of languages $P$ and finetuned on downstream task instances in the language $l \in P$, and $\taskeval{}(M, l)$ is an evaluation of  model $M$ on instances in language $l$ in terms of the downstream metric, e.g., word-label accuracy for part of speech tagging.

Following, we extend the definition of zero-shot transfer score to a set of downstream test languages $D \subseteq P$ to measure  $P$'s aggregated effect on zero-shot performance, by averaging over all bilingual transfer combinations in $D$:

\begin{equation}
\label{eq:zp}
\zs_{P}(D) = \frac{1}{|D|^2 - |D|}\cdot \sum\limits_{\substack{l_1, l_2 \in D \\ l_1 \neq l_2}}\clt{l_1}{l_2}{P}
\end{equation}

In the following sections, we will use these metrics to evaluate how different choices for pretraining languages influence downstream performance.

\section{Data Selection}
\label{sec:data}
We collect a pretraining dataset to test the effect of pretraining languages on cross-lingual transfer.

First, we choose a set of \numoflangs{} languages from 9 language families, as listed in Table~\ref{tab:lang}. These represent a wide variety of scripts, as well as typological and morphological features. We note that our approach can be readily extended to other languages beyond those selected in this study. 

Second, we aim to balance the amount of data and control for its domain across languages, to mitigate possible confounders in our evaluations.
Below we outline design choices we make toward this goal.



\subsection{Data Balancing}
\label{sec:data-balance}
To achieve a balanced dataset across our languages, we sample consecutive sentences from every language's Wikipedia dump from November 2021, such that each language is represented by 10 million characters.\footnote{Wikipedia dump was obtained and cleaned using wikiextractor~\citep{Wikiextractor2015}.} This amount was chosen to align all languages to the lower-resource ones (e.g., Piedmontese or Irish)  which comprise approximately of 10mb. We choose to sample texts from Wikipedia as it consists of roughly similar encyclopedic domain across languages, and is widely used for training PLMs~\citep{devlin-etal-2019-bert}. 



\begin{table}[tb!]
\centering
\small
\resizebox{\columnwidth}{!}{\begin{tabular}{@{}lllcc@{}}
\toprule
\multirow{2}{*}{\textbf{Language}} & \multirow{2}{*}{\textbf{Code}} & \multirow{2}{*}{\textbf{Family}} & \multicolumn{2}{c}{\textbf{Size [M chars]}} \\ 
                                   &                                &                                  & \multicolumn{1}{l}{Wiki} & Sample \\\midrule
Piedmontese       & pms           & Indoeuropean    & 14 & 10                          \\
Irish             & ga            & Indoeuropean    & 38 & 10                         \\
Nepali            & ne            & Indoeuropean    & 78 & 10                          \\
Welsh             & cy            & Indoeuropean    & 85 & 10                          \\
Finnish           & fi            & Uralic          & 131 & 10                        \\
Armenian          & hy            & Indoeuropean    & 174 & 10                        \\
Burmese           & my            & Sino-Tibetian   & 229 & 10                        \\
Hindi             & hi            & Indoeuropean    & 473 & 10                        \\
Telugu            & te            & Dravidian       & 533 & 10                        \\
Tamil             & ta            & Dravidian       & 573 & 10                        \\
Korean            & ko            & Korean          & 756 & 10                        \\
Greek             & el            & Indoeuropean    & 906 & 10                        \\
Hungarian         & hu            & Uralic          & 962 & 10                        \\
Hebrew            & he            & Afroasiatic     & 1,261 & 10                      \\
Chinese           & zh            & Sino-Tibetian   & 1,546  & 10                     \\
Arabic            & ar            & Afroasiatic     & 1,695  & 10                     \\
Swedish            & sv            & Indoeuropean    & 1,744  & 10                     \\
Japanese          & ja            & Japonese        & 3,288  & 10                     \\
French            & fr            & Indoeuropean    & 4,958  & 10                     \\
German            & de            & Indoeuropean    & 6,141  & 10                     \\ 
Russian           & ru            & Indoeuropean    & 6,467  & 10                     \\
English           & en            & Indoeuropean    & 14,433  & 10                    \\\bottomrule
\end{tabular}}
\caption{The size of the full Wikipedia dump for the languages in our study (in millions of characters) versus our fixed sized sampling of it. This exemplifies both the linguistic diversity as well as the variance in data sizes in the original Wikipedia corpus, often used for pretraining PLMs. In contrast, we create a balanced pretraining dataset by sampling 10M characters from all languages such that they conform to the smallest language portion in our set (Piedmontese).}
\label{tab:lang}
\end{table}

\paragraph{Can we balance the amount of information across languages?}
We note that a possible confound in our study is that languages may encode different amounts of \emph{information} in texts of similar character count. This may happen due to differences in the underlying texts or in inherent language properties.\footnote{For example logographic or abjad writing systems may be more condensed than other scripts~\citep{perfetti2005orthography}.}
To estimate the amount of information in each of our $10^7$ character partitions, we tokenize each language partition $l$ with the same word-piece tokenizer, and look at the ratio between the total number of tokens in $l$ and the number of unique tokens in $l$, finding a good correlation across all our languages ($r = 0.73$), which may indicate that our dataset is indeed balanced in terms of information. Our intuition is that an imbalanced amount of information would lead the tokenizer to ``invest'' more tokens in some of the languages while neglecting the less informative ones.

\paragraph{Is our sample representative of the full Wikipedia corpus in each language?}
Another concern may be that our sampled corpus per language is not indicative of the full corpus for that language, which may be much larger (see Table~\ref{tab:lang}). To test this, we create three discrete length distributions. Two length distributions for sentences~(in terms of words and tokens), and word length distribution in terms of characters. We then compare those three distributions between our sample and the full data using Earth Movers Distance. All means and standard deviations score below 0.001, indicating that indeed all samples are similarly distributed to their respective full corpus in terms of these metrics.




\section{Bilingual Pretraining Graph}
\label{sec:language-graph}
\begin{figure*}[tb!]
    \centering
    \includegraphics[width=0.7\textwidth]{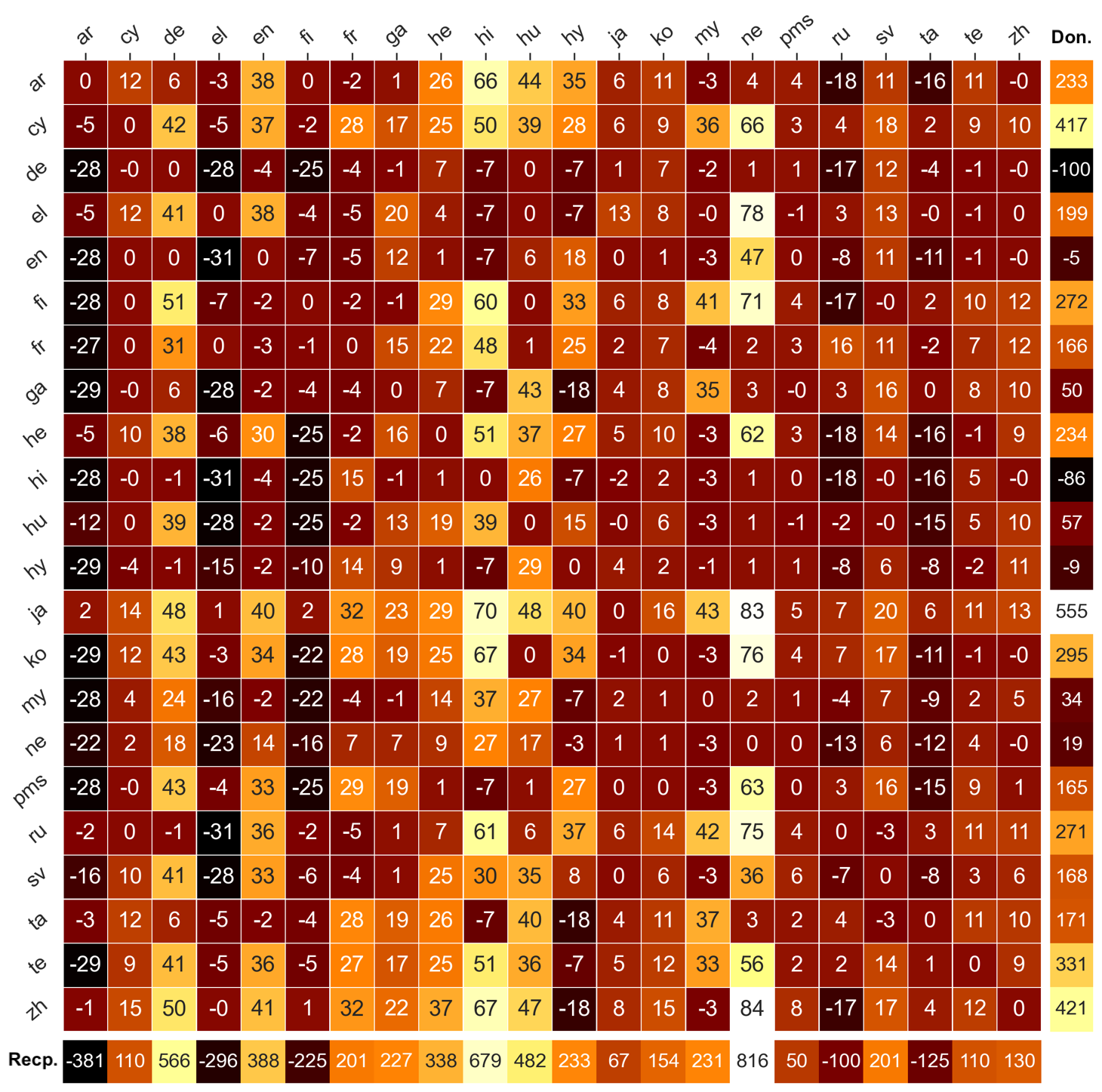}
    
    \caption{Bilingual finetune scores between language pairs in our \emph{balanced} corpus. Coordinate $(i, j)$ represents $\ftt{l_i}{l_j}$, i.e., the performance in MRR[\%] (which correlates with perplexity) of an LM pretrained on a bilingual corpus over languages ($l_i, l_j$) and tested intrinsically on $l_j$. 
    The last column (marked \emph{Don.}) sums over each line, i.e., index $i$ in the column represents how much language $i$ \emph{donated} to all other languages.
    Similarly, the $j$'th index in the last row (marked \emph{Recp.}) sums over column $j$ and represents how much language $l_j$ improved in all configurations.\label{fig:mlm}}
    
\end{figure*}

In this section, we describe a method for estimating the effect that different pretrain language combinations would have on downstream zero-shot performance. This is achieved by evaluating bilingual performance on the pretraining masked language modeling (MLM) task.

We begin by describing our experimental setup, hyperparameters and hardware configuration~(Section \ref{sec:exp-setup}).
In Section~\ref{subsec:language-graph}, we outline our estimation method, yielding a complete graph structure over our languages, amenable for future exploration and analyses (Figures \ref{fig:schwartz}, \ref{fig:mlm}). In the following sections, we use the graph to formulate a set of downstream cross-lingual hypotheses regarding how different languages will affect zero-shot performance, and validate these hypotheses on two downstream tasks.

\subsection{Experimental Setup}
\label{sec:exp-setup}
For all evaluations discussed below, we train a BERT model~\citep{devlin-etal-2019-bert} with 4 layers and 4 attention heads, an MLM task head, and an embedding dimension of 512.\footnote{We use the implementation provided by Hugging Face: \url{https://huggingface.co/bert-base-uncased}.} 
We train a single wordpiece tokenizer~\citep{wu2016google} on our entire dataset.\footnote{  To allow future exploration, we also tokenize over 22 additional languages (listed in the Appendix) which are sampled in the same manner but are not included in this study.}
We train the models with a batch size of 8 samples, with sentences truncated to 128 tokens. 

Each language model was trained up to 4 epochs.
This was determined by examining the training loss on 6 diverse languages in our set and observing that they converge around 4 epochs. A subset of 6 languages was  trained on 4 additional seeds to verify the stability of the results, as seen in Table \ref{tab:pretrain-mean-seeds} and Table  \ref{tab:pretrain-std-seeds} in the Appendix.
Masks were applied with default settings, generating 15\% mask tokens and 10\% random tokens for each input sequence~\citep{devlin-etal-2019-bert}. 
We used a single GPU core (nvidia tesla M60, gtx 980, and RTX 2080Ti). Training time varied between 80 - 120 minutes.



\subsection{Building a Pretraining Language Graph} 
\label{subsec:language-graph}
Intuitively, we measure MLM performance when pretraining on a pair of languages $(l_1, l_2)$ as a proxy to the extent of how $l_1$ and $l_2$ contribute to one another in zero-shot cross-lingual transfer. 

This methodology relies on two assumptions.
First, we assume that the cross-lingual zero-shot performance as defined in Equation~\ref{eq:zp} is \emph{monotonic}, i.e., that adding pretraining languages will improve the average downstream performance. This is defined formally as:
\begin{equation}
    P' \subseteq P \;\; \Rightarrow \;\; \zs_{P'}(D) \leq \zs_{P}(D) 
\end{equation}

Following this assumption will allow us to extend our bilingual observations to a pretraining language set $P$ of arbitrary size.

Second, we assume that MLM performance correlates with downstream task performance, which is often the assumption made when training PLMs to minimize perplexity~\citep{elmo,devlin-etal-2019-bert}. 

\paragraph{Bilingual MLM finetune score.}
Formally, for every language pair $s, t \in P$, we compute the following finetune score, $\ft$: 
\begin{equation}
\label{eq:ftt}
    \ftt{s}{t} := \frac{\taskeval(M^{s, t}, t) - \taskeval(M^{t}, t)}{\taskeval(M^{t}, t)}
\end{equation}

Where $M^{s, t}$ is a model pretrained on $s, t$, and $\taskeval$ is an intrinsic evaluation metric for MLM.\footnote{We specifically use mean reciprocal rank (MRR), which correlates with perplexity.}
I.e., $\ft(s, t)$ estimates how much the target language $t$ ``gains'' in the MLM task from additional pretraining on the source language $s$ compared to monolingual pretraining on $t$.

Figure~\ref{fig:mlm} depicts a weighted adjacency matrix where coordinate $(i, j)$ corresponds to $\ftt{l_i}{l_j}$. As shown in Figure~\ref{fig:schwartz}, the same information can be conveyed in a complete directed weighted graph, where each node represents a language, and edges $(l_1, l_2)$ are weighted by $\ftt{l_1}{l_2}$. 

\paragraph{Language-Level donation and recipience.}
Next, for each language $l \in P$ we compute a \emph{Donation} score, $\donation{}$, as an aggregate over all of its finetune scores as a source language (i.e., how much it contributed to other languages), and similarly a \emph{recipience} score, $\recipience{}$, by aggregating over all its finetune scores as a target language, to measure how much $l$ is contributed to by other languages. Formally:
\begin{align}
    \label{eq:donation}\donation(l) &:= \sum\limits_{\substack{t \in P\\ t \neq l}}\ftt{l}{t} \\
    \label{eq:recipience}\recipience(l) &:= \sum\limits_{\substack{s \in P\\ s \neq l}}\ftt{s}{l}
\end{align}

We depict both donation and recipience scores as aggregate row and column vectors in Figure~\ref{fig:mlm}.

Thus, based on the two assumptions above, our hypothesis is that the downstream cross-lingual transfer will be proportional to the sum of recipience scores for all pretraining languages. Formally:
\begin{equation}
\label{eq:recipience-assumption}
    \zs_{P}(D) \propto \sum\limits_{l \in D} \recipience(l)
\end{equation}
Moreover, higher donation scores for languages in the pretrain set will result in higher scores in the downstream task. Formally:
\begin{equation}
\label{eq:donation-assumption}
    \sum\limits_{l \in P}\donation(l)\leq\sum\limits_{l \in P'}\donation(l) \Rightarrow \zs_{P}(D) \leq \zs_{P'}(D)
\end{equation}

\section{Pretraining Graph Analysis}
\label{sec:pretraining-analysis}
\begin{figure}[tb!]
    \centering
     \includegraphics[width=0.48\textwidth]{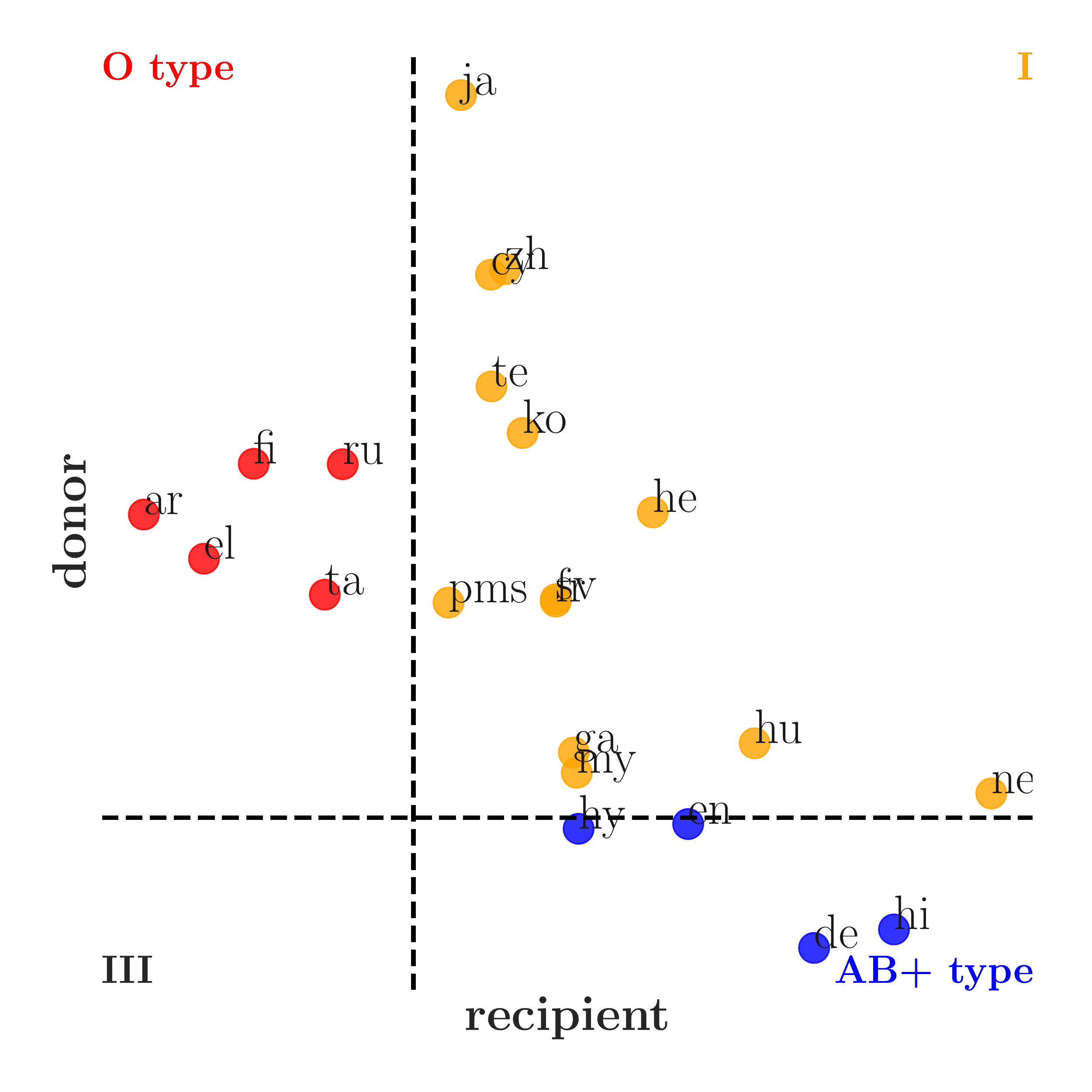}
    \caption{
    Our languages on a ``donor'' versus ``recipient'' axes. A positive coordinate on the ``donor'' score (X axis) represents a language that on average improved other languages' performance in bilingual pretraining, while a negative score indicates a language which hurts other languages on average. Inversely, a positive score on the Y axis represents languages whose performance was improved by bilingual pretraining, while negative scores represent languages whose performance was hurt by it. The $II$ quadrant represents O type languages (donating but not receiving), languages on the $IV$'s quadrant are AB+ type languages (receiving but not donating) 
    \label{fig:lang-type}}
\end{figure}

We present several key observations based on the bilingual pretraining graph described in the previous section and summarized by the adjacency matrix in Figure \ref{fig:mlm}, as well as an interactive exploration interface. 
In the following sections, we use these observations in our downstream evaluations.



\begin{figure*}[tb!]
    \centering
    \includegraphics[width=\linewidth]{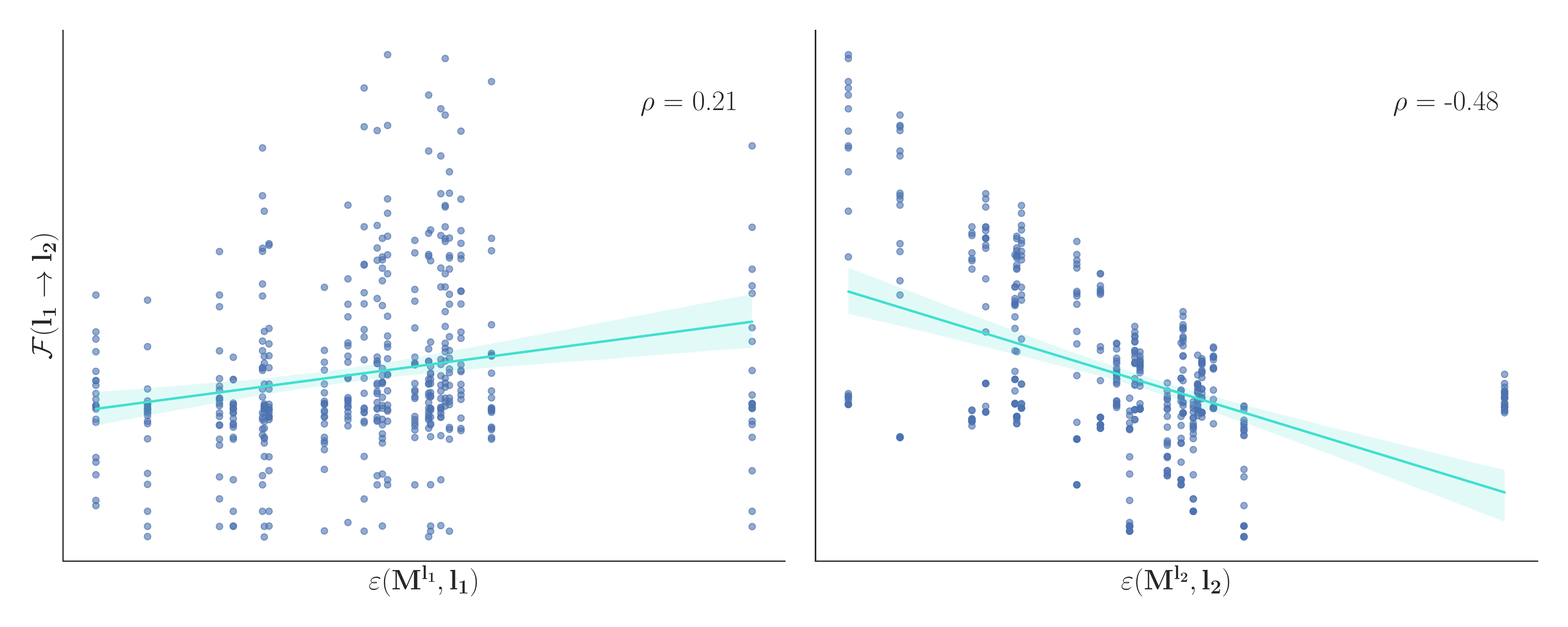}
    \caption{Scatter-plot. Y-axis represents cross-lingual transfer $\ftt{l_1}{l_2}$ for a each possible pair of languages,  
    while the x-axis represents the
    monolingual MRR score for a source language (left) and the target language (right). 
    }
    \label{fig:transfer_correlation}
\end{figure*}

\paragraph{Some language combinations are detrimental.}
Negative finetune scores are present in some of the target languages, e.g., between Korean (ko) and Arabic (ar), which means that initializing a language model for Arabic  with weights learned for Korean hurts MLM performance on Arabic, compared to an Arabic monolingual baseline. I.e., in these language configurations, initializing the model with another language model's weights leads to worse performance than random initialization. 

\paragraph{Bilingual MLM relations are \emph{not} symmetric.}
In fact, we observe a moderate \emph{negative} correlation between $\ftt{l_1}{l_2}$ and $\ftt{l_2}{l_1}$, as shown in Figure~\ref{fig:lang-type}. 
For example, for German and Finnish we get  $0.51=\ftt{fi}{de} > \ftt{de}{fi}=-0.24$. I.e., Finnish initialization improves German MLM, while the inverse  is detrimental for Finnish. 






\paragraph{Monolingual performance correlates with donation score.}
Perhaps expectedly, relatively worse-performing models benefit most from the bilingual transfer, while better-performing monolingual models tend to be better donors, although to a lesser extent (Figure~\ref{fig:transfer_correlation}).\footnote{Correlations are statistically significant ($p<0.05$ based on Student's T-test).}

\begin{figure}[tb!]
    \centering
    \begin{subfigure}[b]{0.45\linewidth}
        \includegraphics[width=\linewidth]{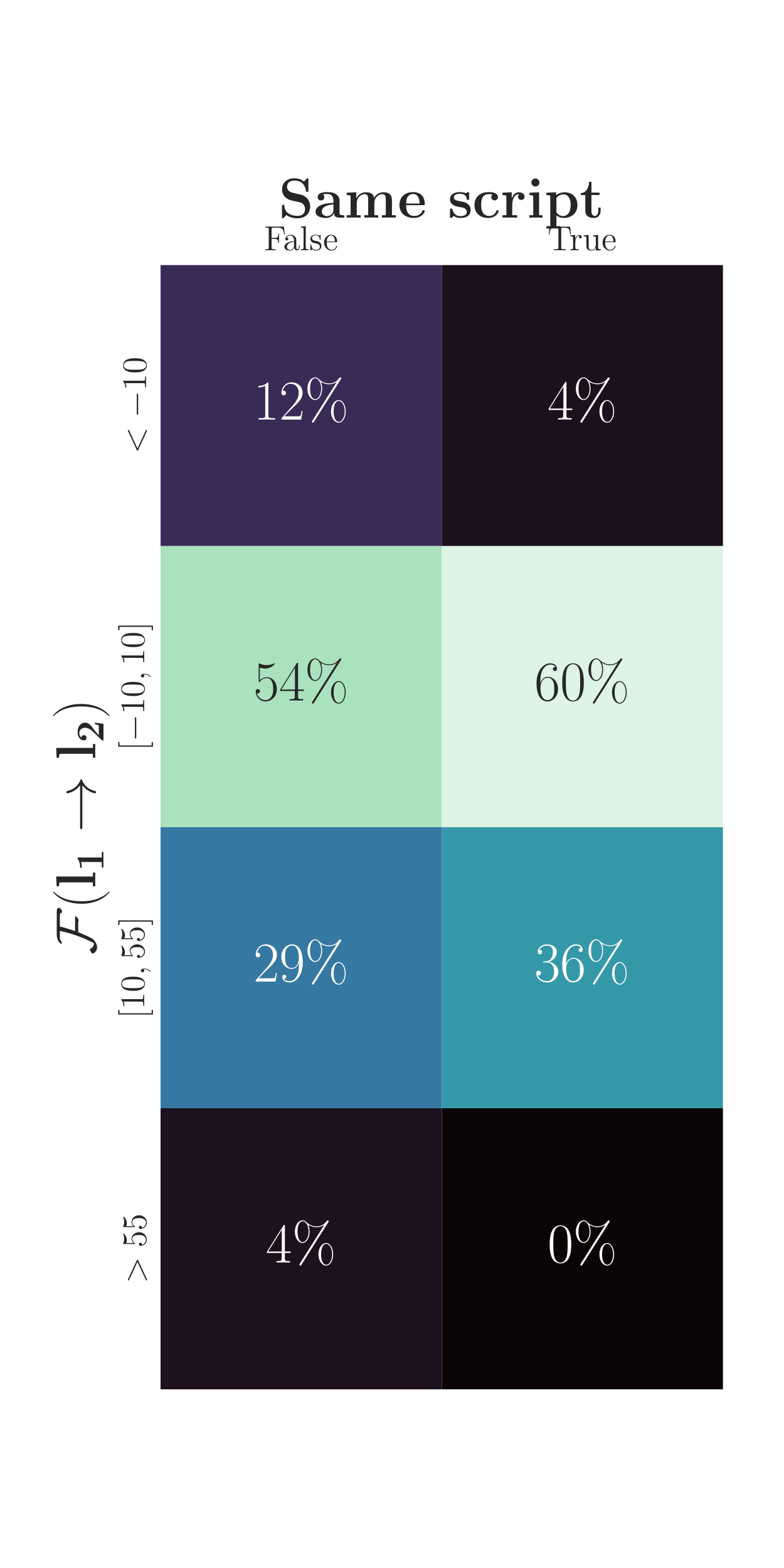}
        \caption{Sharing Script}
        \label{fig:cross_script}
    \end{subfigure}%
    \hfill%
    \begin{subfigure}[b]{0.45\linewidth}
        \includegraphics[width=\linewidth]{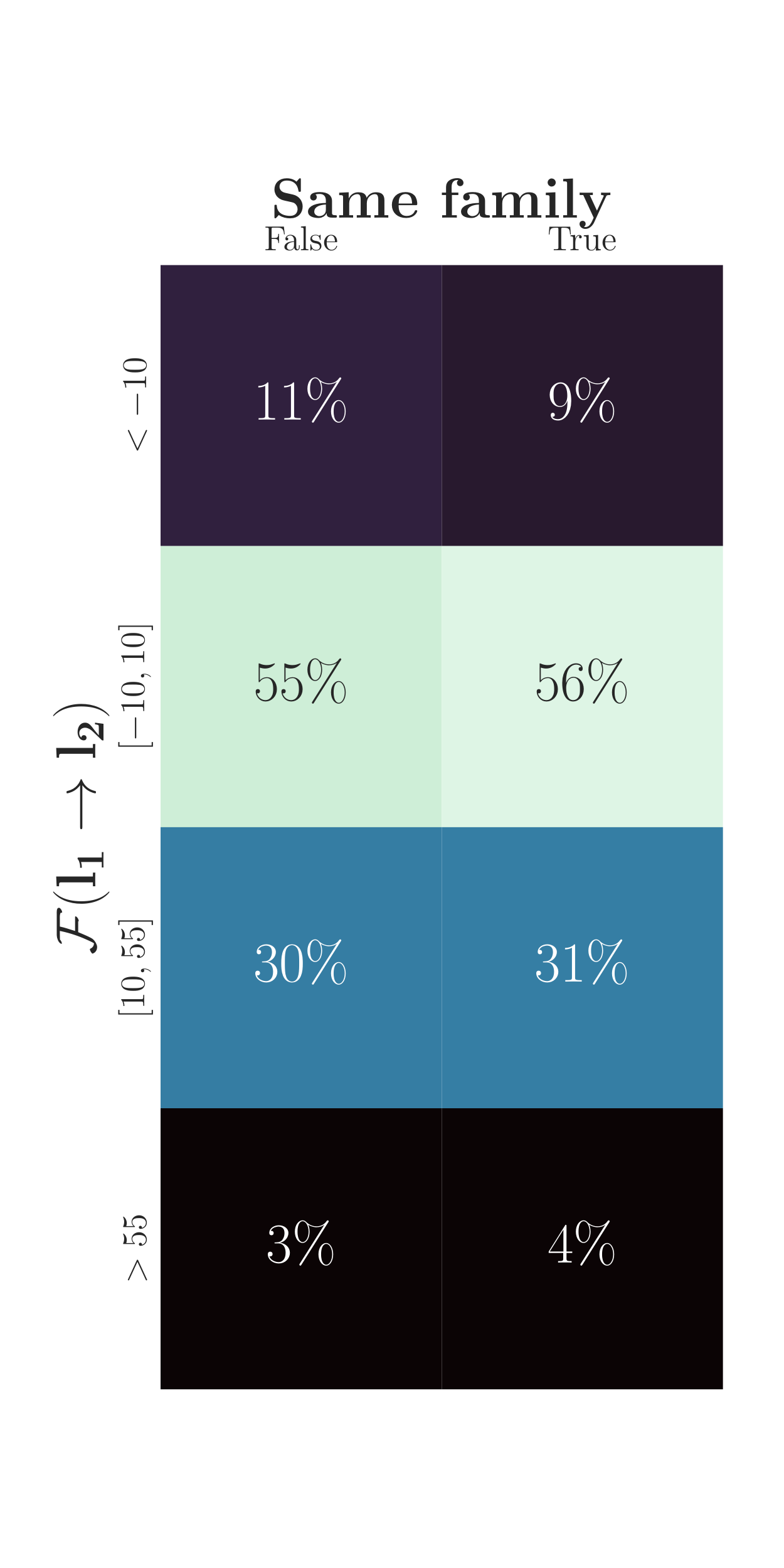}
        \caption{Sharing Family}
        \label{fig:cross_family}
    \end{subfigure}
    \caption{We divide language pairs into four bins by bilingual finetune score ($\ftt{l_1}{l_2}$).\footnotemark 
    The figures present the percentage of pairs assigned to each bin for samples of language pairs: (a) written in the same or different script; (b) belonging to the same or different language family. 
    Sharing the language family has no significant effect on the transfer score $(p>0.05)$, while the effect of sharing scripts is significant $(p<0.05)$  (p-values based on Pearson's $\chi^2$ test). }
\end{figure}
\footnotetext{We motivate our choice of bins in Appendix.}




\paragraph{Different script leads to larger variance in bilingual finetuning. However, language family does not affect it.}
We find that fine-tuning between languages with different scripts is a high-risk, high-reward scenario. The highest transfer scores occur in this setting, but the proportion of negative scores is also higher.
A shared script is a safe setting with a high proportion of neutral or positive donations~(Figure~\ref{fig:cross_script}).
In contrast with recent findings~\cite{pires-etal-2019-multilingual}, we did not observe a statistically significant influence for the language family~(Figure~\ref{fig:cross_family}).




\paragraph{Finetuning as transfusion: mapping the linguistic blood-bank.}
The non-symmetric nature of the scores gives rise to a coarse-grained ontology loosely reminiscent of human blood types, depicted in Figure~\ref{fig:lang-type}. 
Languages which on average donate but do not receive ($\donation(l) > 0$ and $\recipience(l) < 0$) are denoted \emph{O type languages}, while the inverse (receiving but not donating) are denoted as \emph{AB+ type}.

\subsection{Interactive Exploration}
\label{sec:vis_tool}
To allow further exploration of our bilingual pretraining graph, we develop a publicly available web-based interactive exploration interface.\footnote{\demo{}}
We enable exploration of interactions between different linguistic features, based on \emph{The World Atlas of Language Structures} (WALS)~\citep{wals},  allowing users to filter and focus on specific traits and analyze  how they affect bilingual pretraining.

\section{Downstream Zero-Shot Performance}
\label{sec:downstream}
In this section, we validate our method for estimating the effect of pretraining language combinations on downstream performance.
Towards that end, in Section~\ref{sec:choosing-sets}, we construct several pretraining configurations, based on pretraining observations.
Then, in Section \ref{sec:ds-tasks} we describe the multilingual datasets we use for two downstream tasks.
Finally, our results are presented in Section \ref{sec:downstream-results}, showing the influence of pretraining configuration on downstream performance.


\subsection{Choosing Pretraining Sets}
\label{sec:choosing-sets}
We use the donation scores to identify pretraining languages projected to lead to better downstream zero-shot performance, and the recipience score to find downstream languages which will perform better languages as source (finetune) languages. Our setup is summarized in Table~\ref{tab:ds-choices}.

\paragraph{Donating languages.}
We define three sets of languages for pretraining, using the donation score while keeping the sets linguistically diverse: (1) \emph{Most Donating:} Japanese, Telugu, Finnish, and Russian;
(2) \emph{Least Donating:} Nepali, Burmese, Armenian, and English. We also include Englishs as it is a popular source language; 
and (3) \emph{Random:}A randomly selected set of 4 languages: Hebrew, Irish, French and Swedish. 

\paragraph{Recipient languages.}
To validate that lower recipience scores indeed indicate that languages are less likely to improve via cross lingual transfer, we added 6 languages to all configurations described above: 3 \emph{Most Recipient} languages ($R_h$): Hindi, German, and Hungarian, and 3 \emph{Least Recipient} languages ($R_l$): Arabic, Greek, and Tamil. Finally, we add a fourth control configuration which was pretrained only on $C:=R_h\cup{}R_l$.

\paragraph{Hypotheses.}
We hypothesize that the more donating pretraining sets will improve cross-lingual transfer in downstream tasks, and that more recipient languages will have better cross-lingual performance compared to least recipient languages. These can be formally articulated using Equations \ref{eq:recipience-assumption-exp} and \ref{eq:donation-assumption-exp}: 
 
\begin{equation}
\label{eq:recipience-assumption-exp}
    \forall P:\zs_{P}(R_h)>\zs_{P}(R_l)
\end{equation}
\begin{equation}
\label{eq:donation-assumption-exp}
    \zs_{Most Don.}(C) > \zs_{Random}(C) > \zs_{Least Don.}(C)
\end{equation}

\subsection{Tasks}
\label{sec:ds-tasks}
We evaluated all pretraining configurations detailed in Table~\ref{tab:ds-choices} on two of \xtreme{}'s tasks: part of speech tagging (POS) and named entity recognition (NER). Both of which commonly appear in NLP pipelines such as CoreNLP~\citep{manning-etal-2014-stanford} and spaCy~\citep{spacy2}.
We aim to balance the data in both tasks across different finetune languages, so as not to skew results towards higher-resource languages. 

For part-of-speech tagging, \xtreme{} borrows from universal dependencies~\citep{nivre2020universal}. 
Since \xtreme{} is imbalanced across languages,
we truncated the data to 1000 sentences to fit the lower-resource languages, e.g., \xtreme{} annotates POS in 909 sentences in Hungarian. 
For NER, we applied a similar procedure, where \xtreme{}'s data was taken from the Wikiann (panx) dataset~\citep{rahimi2019massively} which we truncated to 5000 sentences (the data size available for Hindi NER in \xtreme{}).

\paragraph{Experimental setup.} 
We use the code and default hyperparameter default values provided by \xtreme{} to train the downstream tasks~\citep{hu2020xtreme}, adapted for multilingual training.


\begin{table*}[tb!]
\small
\centering
\resizebox{\textwidth}{!}{\begin{tabular}{lclccccl}
\toprule
 &
  \multicolumn{2}{c}{\textbf{Base Pretrain Set}} &
  \multicolumn{3}{c}{\textbf{Shared Pretrain Set}} &
  \textbf{Total Data} &
  \multicolumn{1}{c}{\textbf{Summary}} \\
 &
  \multicolumn{2}{c}{\textbf{(Donors)}} &
  \textbf{Most Recipient ($R_h$)} &
   &
  \textbf{Least Recipient ($R_l$)} &
   &
  \multicolumn{1}{c}{} \\ \cmidrule{2-8} 
\textbf{Most Donating} &
  \{ja, te, fi, ru\} &
  \multirow{4}{*}{+} &
  \{hi, de, hu\} &
  \multirow{4}{*}{+} &
  \{ar, el, ta\} &
  $10^8$ characters &
  Most donating pretraining set. \\
\textbf{Least Donating} &
  \{ne, my, hy, en\} &
   &
  \{hi, de, hu\} &
   &
  \{ar, el, ta\} &
  $10^8$ characters &
  Least donating pretrain set. \\
\textbf{Random} &
  \{he, ga, fr, sv\} &
   &
  \{hi, de, hu\} &
   &
  \{ar, el, ta\} &
  $10^8$ characters &
  Random donating pretrain set. \\
\textbf{Control} &
  \{\} &
   &
  \{hi, de, hu\} &
   &
  \{ar, el, ta\} &
  $10^8$ characters &
  No additional donating languages. \\ \bottomrule
\end{tabular}}
\caption{Four pretraining language configurations. Each consists of \emph{donating} languages (first column) and \emph{recipient} languages (second column).
The control group has the same amount of data, equally distributed among its languages.}
\label{tab:ds-choices}
\end{table*}


\begin{table*}[tb!]
\small
\centering
\begin{tabular}{@{}lcccc@{}}
\toprule
\multirow{2}{*}{} & \multicolumn{2}{c}{\textbf{NER} [$\%F_1$]} & \multicolumn{2}{c}{\textbf{POS} [$\%F_1$]} \\
                  & Avg. Monolingual     & Avg. Zeroshot       & Avg. Monolingual        & Avg. Zeroshot    \\ \cmidrule(l){2-5} 
\textbf{Most Donating}      & \textbf{49.3$\pm.4$}  & \textbf{15.6$\pm.4$} & \textbf{61.4$\pm.1$}              & \textbf{28.1$\pm.3$}        \\
\textbf{Random}   & \textbf{49.2$\pm.3$}  & \textbf{15.6$\pm.1$} & 61.3$\pm.1$              & 26.9$\pm.3$        \\
\textbf{Least Donating}   & 48.8$\pm.2$           & 14.8$\pm.3$          & 60.9$\pm.2$              & 26.9$\pm.6$        \\ \cmidrule(l){2-5} 
\textbf{Control}  & 49.0$\pm.2$           & \textbf{15.6$\pm.2$}          & \textbf{61.9$\pm.1$}     & 27.4$\pm.3$       \\ \bottomrule
\end{tabular}
\caption{Donation results for named entity recognition (NER) and part of speech tagging (POS) as mean and standard deviation over five random seeds. For each pretraining language group (\emph{Most Donating}, \emph{Random}, \emph{Least Donating}, and \emph{Control}), we report corresponding average monolingual and zero shot performance. \emph{Most Donating} consistently outperforms \emph{Least Donating} in both tasks, and in both monolingual and zeroshot performance. \emph{Most Donating}  is on par with  \emph{Control} in monolingual performance in NER, despite having less in-domain data.}

\label{tab:inf-results}
\end{table*}

\begin{table*}[tb!]
\small
\centering
\begin{tabular}{@{}lcccc@{}}
\toprule
\multirow{2}{*}{}                                    & \multicolumn{2}{c}{\textbf{NER} [$\%F_1$]} & \multicolumn{2}{c}{\textbf{POS} [$\%F_1$]} \\
                                                     & Avg. Monolingual      & Avg. Zeroshot      & Avg. Monolingual      & Avg. Zeroshot      \\ \cmidrule(l){2-5} 
\multicolumn{1}{c}{\textbf{Most Recipient ($R_h$)}} &\textbf{50.3$\pm.6$} & \textbf{18.4$\pm.6$} & \textbf{64.1$\pm.3$} & \textbf{28.7$\pm.7$} \\
\multicolumn{1}{c}{\textbf{Least Recipient ($R_l$)}} & 47.9$\pm.4$        &    12.4$\pm.4$         & 58.6$\pm.4$           & 26.0$\pm.7$         \\ \bottomrule
\end{tabular}
\caption{Recipience results for named entity recognition (NER) and part of speech tagging (POS) as mean and standard deviation over five random seeds. We report results across different training configurations for two groups of downstream recipient languages. 
In accordance with our pretraining results, the \emph{Most Recipient} set does better than the \emph{Least recipient} set across both tasks in zero-shot and monolingual performance.
\label{tab:cont-results}}
\end{table*}

\subsection{Results}
\label{sec:downstream-results}
Several key observations can be made based on the results for both POS tagging and NER across all training configurations, which are presented in Tables \ref{tab:inf-results} and \ref{tab:cont-results}. 
For each configuration $P$ in \emph{Most Donating, Least Donating, Random, Control} we calculated zero-shot transfer scores on $C$, using $\zs_{P}(C)$ defined by Equation \ref{eq:zp}. Monolingual results under each pretrain set $P$ were calculated by the average $F1$ performance of each language in $C$:
\begin{equation}
    \frac{1}{|C|} \cdot \sum\limits_{{l \in C}}\taskeval{}(M^{P,l}, l)
\end{equation}
Where $\taskeval{}(M^{P,l}, l)$ denotes the $F1$ score of a model pretrained on $P$, finetuned on $l$ and evaluated on $l$.

\paragraph{Pretraining configuration affects downstream cross-lingual transfer.}
In both tasks, we observe a variance in results when changing the pretraining configuration, despite all of them having similar amounts of data. 
This may imply that previous work  has  omitted an important interfering factor.

\paragraph{Recipience score correlates with downstream cross-lingual performance.} 
We evaluated zero-shot transfer for each language set $R \in \{R_l, R_h$\} as the average zero-shot transfer scores over all pretraining configurations.
Table \ref{tab:cont-results} reveals that the \emph{Most Recipient} set outperforms the \emph{Least Recipient} set in both tasks~($+5.5\%$ in NER, $+2.7\%$ in POS tagging). 

\paragraph{Multilingual pretraining can improve \emph{monolingual} performance.} As seen in Table \ref{tab:inf-results}, the \emph{Most Donating} pretraining configuration achieved a monolingual score which is slightly higher than the control group, while the \emph{Least Donating} configuration underperforms all other sets. This suggests that multilingual pretraining datasets can benefit monolingual downstream results compared to more data in a single language. 

\paragraph{English might not be an optimal pretraining language.} Corresponding with our previous results, if donation score is indicative of a language's contribution in pretraining, English's relative low donation score might indicate that it is not the best language to pretrain upon. English was also part of the \emph{Least Donating} pretraining configuration which scored lower than \emph{Most Donating} as seen in Table \ref{tab:inf-results}. Further research can ascertain this finding.



\section{Limitations and Future Work}
As with other works on cross-lingual transfer, our results are influenced by many hyperparameters.
Below we explicitly define our design choices and how they can be explored in future work.

First, data scarcity in low-resource languages restricted us to small data amounts. Although our experiments showed a non-trivial signal for pretraining and downstream tasks, 
future work may apply our framework to larger data sizes.

Second, for efficiency's sake, we trained relatively small models to enable us to train a large number of language configurations, while ensuring convergence in 6 languages.  
Furthermore, we did not do any hyper-parameter tuning and used only values reported in previous work, and use only the BERT architecture. Future work may revisit any of these design choices to shed more light on their effect.

Third, similarly to other works, our data was scraped from Wikipedia, and we did not account for language contamination across supposedly monolingual corpora (e.g., due to code switching). Such contamination may confound with cross-lingual transfer, as was recently shown by \citet{blevins2022language}.

Finally, our downstream analysis focused on POS tagging and NER since they were available for many languages and are common in many NLP pipelines. Further experimentation can test if our results hold for more NLP tasks.

\section{Related Work}
\label{sec:related}
To the best of our knowledge, this is the first work to control for the amount of data allocated for each language during pretraining and finetuning while evaluating on many languages. 

Perhaps most related to our work, \citet{turc2021revisiting} challenge the primacy of English as a source language for cross-lingual transfer in various downstream tasks. Their work shows that German and Russian are often more effective sources. In all of their experiments, they use \mbert{}'s imbalanced pretraining corpus. \citet{blevins2022language} complement this hypothesis by showing that English pretraining data actually contains a significant amount of non-English text, which correlates with the model's transfer capabilities.

\citet{wu2020all} evaluate how \mbert{} performs on a wide set of languages, focusing on the quality of representation for low-resource languages in various downstream tasks by defining a scale from low to high resource. They show that \mbert{} underperforms non BERT monolingual baselines for low resource languages while performing well for high resource ones. 

While \citet{pires-etal-2019-multilingual, limisiewicz2021examining} show that typology plays a significant role for \mbert{}'s multilingual performance, this is not replicated in our balanced evaluation, and has lesser impact in \citet{wu2022oolong} as well.

Finally, \citet{conneau2019unsupervised} introduce the transfer-interference trade-off where low resource languages benefit from multilingual training, up to a  point where the overall performance on monolingual and cross-lingual benchmarks degrades.



\section{Conclusions}
\label{sec:conclusions}
We explored the effect of pretraining language selection on downstream zero-shot transfer. 

We first choose a diverse pretraining set of 22 languages, and curate a pretraining corpus which is balanced across these languages.

Second, we devise an estimation technique, quadratic in the number of languages, projecting which pretraining languages will serve better in cross-lingual transfer and which specific downstream languages will do best in that setting. 

Finally, we test our hypothesis on two downstream multilignual tasks, and show that the choice of pretraining languages indeed leads to varying downstream cross-lingual results, and that our method is a good estimation for downstream performance. 
Taken together, our results suggest that pretraining language selection should be a factor in estimating cross-lingual transfer, and that current practices which focus on high-resource languages may be sub-optimal.


\section*{Acknowledgements}
We would like to thank Roy Schwartz for his helpful comments and suggestions and the anonymous reviewers for their valuable feedback. This work was supported in part by a research gift from the Allen Institute for AI. Tomasz Limisiewicz's visit to the Hebrew University has been supported by grant 338521 of the Charles University Grant Agency and the Mobility Fund of Charles University.

\bibliography{custom}

\begin{thebibliography}{22}
\expandafter\ifx\csname natexlab\endcsname\relax\def\natexlab#1{#1}\fi

\bibitem[{Attardi(2015)}]{Wikiextractor2015}
Giusepppe Attardi. 2015.
\newblock Wikiextractor.
\newblock \url{https://github.com/attardi/wikiextractor}.

\bibitem[{Blevins and Zettlemoyer(2022)}]{blevins2022language}
Terra Blevins and Luke Zettlemoyer. 2022.
\newblock \href {https://arxiv.org/abs/2204.08110} {Language contamination
  explains the cross-lingual capabilities of english pretrained models}.
\newblock \emph{ArXiv preprint}, abs/2204.08110.

\bibitem[{Conneau et~al.(2020{\natexlab{a}})Conneau, Khandelwal, Goyal,
  Chaudhary, Wenzek, Guzm{\'a}n, Grave, Ott, Zettlemoyer, and
  Stoyanov}]{conneau2019unsupervised}
Alexis Conneau, Kartikay Khandelwal, Naman Goyal, Vishrav Chaudhary, Guillaume
  Wenzek, Francisco Guzm{\'a}n, Edouard Grave, Myle Ott, Luke Zettlemoyer, and
  Veselin Stoyanov. 2020{\natexlab{a}}.
\newblock \href {https://doi.org/10.18653/v1/2020.acl-main.747} {Unsupervised
  cross-lingual representation learning at scale}.
\newblock In \emph{Proceedings of the 58th Annual Meeting of the Association
  for Computational Linguistics}, pages 8440--8451, Online. Association for
  Computational Linguistics.

\bibitem[{Conneau et~al.(2020{\natexlab{b}})Conneau, Wu, Li, Zettlemoyer, and
  Stoyanov}]{wu2019emerging}
Alexis Conneau, Shijie Wu, Haoran Li, Luke Zettlemoyer, and Veselin Stoyanov.
  2020{\natexlab{b}}.
\newblock \href {https://doi.org/10.18653/v1/2020.acl-main.536} {Emerging
  cross-lingual structure in pretrained language models}.
\newblock In \emph{Proceedings of the 58th Annual Meeting of the Association
  for Computational Linguistics}, pages 6022--6034, Online. Association for
  Computational Linguistics.

\bibitem[{Devlin et~al.(2019)Devlin, Chang, Lee, and
  Toutanova}]{devlin-etal-2019-bert}
Jacob Devlin, Ming-Wei Chang, Kenton Lee, and Kristina Toutanova. 2019.
\newblock \href {https://doi.org/10.18653/v1/N19-1423} {{BERT}: Pre-training of
  deep bidirectional transformers for language understanding}.
\newblock In \emph{Proceedings of the 2019 Conference of the North {A}merican
  Chapter of the Association for Computational Linguistics: Human Language
  Technologies, Volume 1 (Long and Short Papers)}, pages 4171--4186,
  Minneapolis, Minnesota. Association for Computational Linguistics.

\bibitem[{Dryer and Haspelmath(2013)}]{wals}
Matthew~S. Dryer and Martin Haspelmath, editors. 2013.
\newblock \href {https://wals.info/} {\emph{WALS Online}}.
\newblock Max Planck Institute for Evolutionary Anthropology, Leipzig.

\bibitem[{Honnibal and Montani(2017)}]{spacy2}
Matthew Honnibal and Ines Montani. 2017.
\newblock {spaCy 2}: Natural language understanding with {B}loom embeddings,
  convolutional neural networks and incremental parsing.
\newblock To appear.

\bibitem[{Hu et~al.(2020)Hu, Ruder, Siddhant, Neubig, Firat, and
  Johnson}]{hu2020xtreme}
Junjie Hu, Sebastian Ruder, Aditya Siddhant, Graham Neubig, Orhan Firat, and
  Melvin Johnson. 2020.
\newblock \href {http://proceedings.mlr.press/v119/hu20b.html} {{XTREME:} {A}
  massively multilingual multi-task benchmark for evaluating cross-lingual
  generalisation}.
\newblock In \emph{Proceedings of the 37th International Conference on Machine
  Learning, {ICML} 2020, 13-18 July 2020, Virtual Event}, volume 119 of
  \emph{Proceedings of Machine Learning Research}, pages 4411--4421. {PMLR}.

\bibitem[{Joshi et~al.(2020)Joshi, Santy, Budhiraja, Bali, and
  Choudhury}]{joshi-etal-2020-state}
Pratik Joshi, Sebastin Santy, Amar Budhiraja, Kalika Bali, and Monojit
  Choudhury. 2020.
\newblock \href {https://doi.org/10.18653/v1/2020.acl-main.560} {The state and
  fate of linguistic diversity and inclusion in the {NLP} world}.
\newblock In \emph{Proceedings of the 58th Annual Meeting of the Association
  for Computational Linguistics}, pages 6282--6293, Online. Association for
  Computational Linguistics.

\bibitem[{K et~al.(2020)K, Wang, Mayhew, and Roth}]{wang2019cross}
Karthikeyan K, Zihan Wang, Stephen Mayhew, and Dan Roth. 2020.
\newblock \href {https://openreview.net/forum?id=HJeT3yrtDr} {Cross-lingual
  ability of multilingual {BERT:} an empirical study}.
\newblock In \emph{8th International Conference on Learning Representations,
  {ICLR} 2020, Addis Ababa, Ethiopia, April 26-30, 2020}. OpenReview.net.

\bibitem[{Lazar et~al.(2021)Lazar, Saret, Yehudai, Horowitz, Wasserman, and
  Stanovsky}]{lazar-etal-2021-filling}
Koren Lazar, Benny Saret, Asaf Yehudai, Wayne Horowitz, Nathan Wasserman, and
  Gabriel Stanovsky. 2021.
\newblock \href {https://doi.org/10.18653/v1/2021.emnlp-main.384} {Filling the
  gaps in {A}ncient {A}kkadian texts: A masked language modelling approach}.
\newblock In \emph{Proceedings of the 2021 Conference on Empirical Methods in
  Natural Language Processing}, pages 4682--4691, Online and Punta Cana,
  Dominican Republic. Association for Computational Linguistics.

\bibitem[{Limisiewicz and Mare{\v{c}}ek(2021)}]{limisiewicz2021examining}
Tomasz Limisiewicz and David Mare{\v{c}}ek. 2021.
\newblock \href {https://doi.org/10.18653/v1/2021.emnlp-main.376} {Examining
  cross-lingual contextual embeddings with orthogonal structural probes}.
\newblock In \emph{Proceedings of the 2021 Conference on Empirical Methods in
  Natural Language Processing}, pages 4589--4598, Online and Punta Cana,
  Dominican Republic. Association for Computational Linguistics.

\bibitem[{Manning et~al.(2014)Manning, Surdeanu, Bauer, Finkel, Bethard, and
  McClosky}]{manning-etal-2014-stanford}
Christopher Manning, Mihai Surdeanu, John Bauer, Jenny Finkel, Steven Bethard,
  and David McClosky. 2014.
\newblock \href {https://doi.org/10.3115/v1/P14-5010} {The {S}tanford
  {C}ore{NLP} natural language processing toolkit}.
\newblock In \emph{Proceedings of 52nd Annual Meeting of the Association for
  Computational Linguistics: System Demonstrations}, pages 55--60, Baltimore,
  Maryland. Association for Computational Linguistics.

\bibitem[{Nivre et~al.(2020)Nivre, de~Marneffe, Ginter, Haji{\v{c}}, Manning,
  Pyysalo, Schuster, Tyers, and Zeman}]{nivre2020universal}
Joakim Nivre, Marie-Catherine de~Marneffe, Filip Ginter, Jan Haji{\v{c}},
  Christopher~D. Manning, Sampo Pyysalo, Sebastian Schuster, Francis Tyers, and
  Daniel Zeman. 2020.
\newblock \href {https://aclanthology.org/2020.lrec-1.497} {{U}niversal
  {D}ependencies v2: An evergrowing multilingual treebank collection}.
\newblock In \emph{Proceedings of the 12th Language Resources and Evaluation
  Conference}, pages 4034--4043, Marseille, France. European Language Resources
  Association.

\bibitem[{Perfetti and Liu(2005)}]{perfetti2005orthography}
Charles~A Perfetti and Ying Liu. 2005.
\newblock Orthography to phonology and meaning: Comparisons across and within
  writing systems.
\newblock \emph{Reading and Writing}, 18(3):193--210.

\bibitem[{Peters et~al.(2018)Peters, Neumann, Iyyer, Gardner, Clark, Lee, and
  Zettlemoyer}]{elmo}
Matthew~E. Peters, Mark Neumann, Mohit Iyyer, Matt Gardner, Christopher Clark,
  Kenton Lee, and Luke Zettlemoyer. 2018.
\newblock \href {https://doi.org/10.18653/v1/N18-1202} {Deep contextualized
  word representations}.
\newblock In \emph{Proceedings of the 2018 Conference of the North {A}merican
  Chapter of the Association for Computational Linguistics: Human Language
  Technologies, Volume 1 (Long Papers)}, pages 2227--2237, New Orleans,
  Louisiana. Association for Computational Linguistics.

\bibitem[{Pires et~al.(2019)Pires, Schlinger, and
  Garrette}]{pires-etal-2019-multilingual}
Telmo Pires, Eva Schlinger, and Dan Garrette. 2019.
\newblock \href {https://doi.org/10.18653/v1/P19-1493} {How multilingual is
  multilingual {BERT}?}
\newblock In \emph{Proceedings of the 57th Annual Meeting of the Association
  for Computational Linguistics}, pages 4996--5001, Florence, Italy.
  Association for Computational Linguistics.

\bibitem[{Rahimi et~al.(2019)Rahimi, Li, and Cohn}]{rahimi2019massively}
Afshin Rahimi, Yuan Li, and Trevor Cohn. 2019.
\newblock \href {https://doi.org/10.18653/v1/P19-1015} {Massively multilingual
  transfer for {NER}}.
\newblock In \emph{Proceedings of the 57th Annual Meeting of the Association
  for Computational Linguistics}, pages 151--164, Florence, Italy. Association
  for Computational Linguistics.

\bibitem[{Turc et~al.(2021)Turc, Lee, Eisenstein, Chang, and
  Toutanova}]{turc2021revisiting}
Iulia Turc, Kenton Lee, Jacob Eisenstein, Ming-Wei Chang, and Kristina
  Toutanova. 2021.
\newblock \href {https://arxiv.org/abs/2106.16171} {Revisiting the primacy of
  english in zero-shot cross-lingual transfer}.
\newblock \emph{ArXiv preprint}, abs/2106.16171.

\bibitem[{Wu and Dredze(2020)}]{wu2020all}
Shijie Wu and Mark Dredze. 2020.
\newblock \href {https://doi.org/10.18653/v1/2020.repl4nlp-1.16} {Are all
  languages created equal in multilingual {BERT}?}
\newblock In \emph{Proceedings of the 5th Workshop on Representation Learning
  for NLP}, pages 120--130, Online. Association for Computational Linguistics.

\bibitem[{Wu et~al.(2016)Wu, Schuster, Chen, Le, Norouzi, Macherey, Krikun,
  Cao, Gao, Macherey et~al.}]{wu2016google}
Yonghui Wu, Mike Schuster, Zhifeng Chen, Quoc~V Le, Mohammad Norouzi, Wolfgang
  Macherey, Maxim Krikun, Yuan Cao, Qin Gao, Klaus Macherey, et~al. 2016.
\newblock \href {https://arxiv.org/abs/1609.08144} {Google's neural machine
  translation system: Bridging the gap between human and machine translation}.
\newblock \emph{ArXiv preprint}, abs/1609.08144.

\bibitem[{Wu et~al.(2022)Wu, Papadimitriou, and Tamkin}]{wu2022oolong}
Zhengxuan Wu, Isabel Papadimitriou, and Alex Tamkin. 2022.
\newblock \href {https://arxiv.org/abs/2202.12312} {Oolong: Investigating what
  makes crosslingual transfer hard with controlled studies}.
\newblock \emph{ArXiv preprint}, abs/2202.12312.

\end{thebibliography}
\bibliographystyle{acl_natbib}
\clearpage

\appendix
\section{Appendix}
\label{sec:appendix}
\begin{figure}[tb!]
    \centering
    \includegraphics[width=0.7\textwidth]{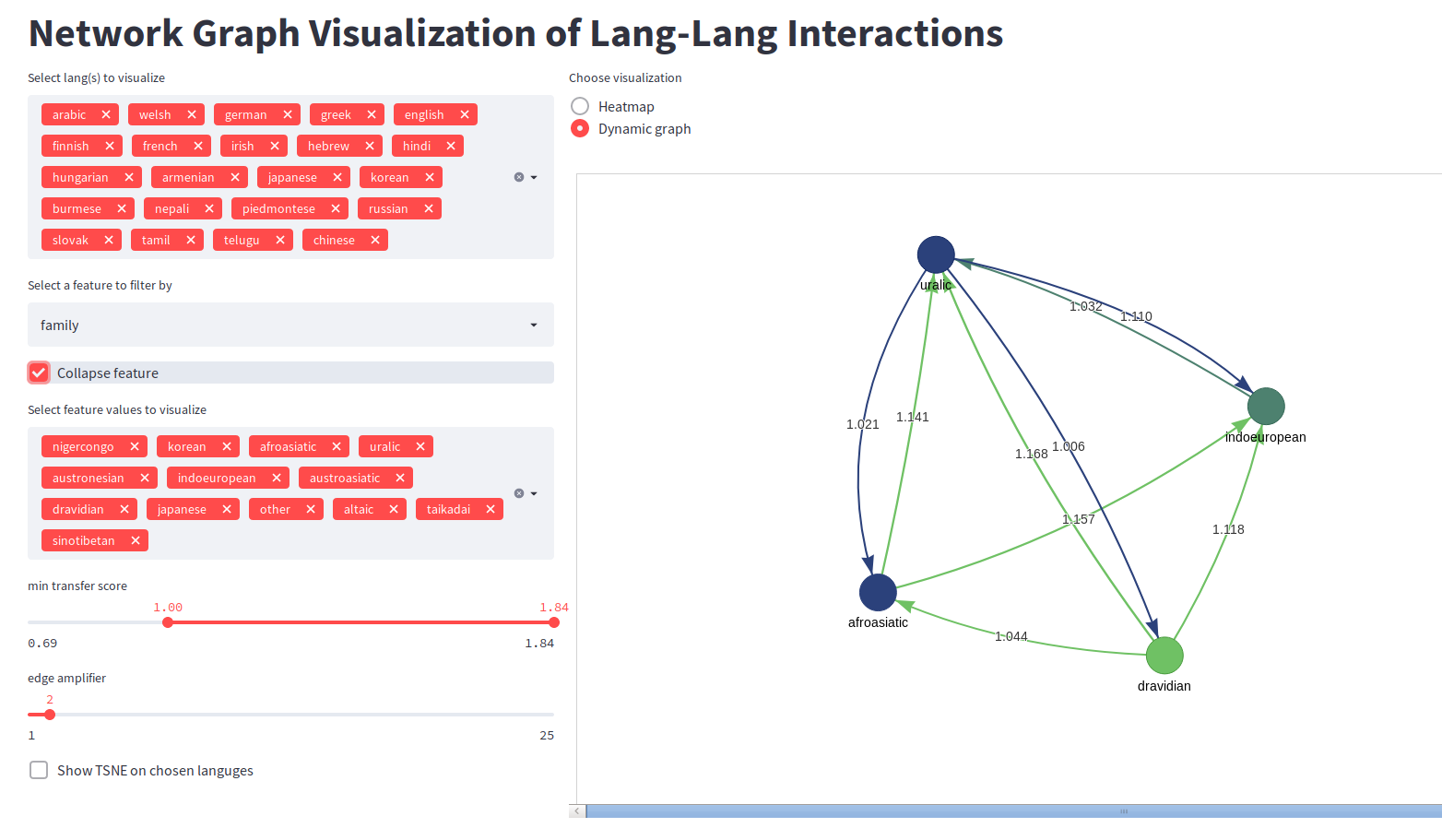}
    \caption{Our visualization tool, based on Streamlit (\url{https://streamlit.io}) 
    \label{fig:tool}}
\end{figure}

\paragraph{Full list of tokenized languages} The full list of Wikipedia language codes for languages used in our tokenizer training is: 

\begin{itemize}
\item pms, ga, ne, cy, fi, hy, my, hi, te, ta, ko, el, hu, he, zh, ar, sv, ja, fr, de, ru, en - languages that are also evaluated and trained. Elaborated in Table \ref{tab:lang}. 

\item af, am, ca, cs, da, es, id, is, it, mg, nl, pl, sk, sw, th, tr, ur, vi, yi - Additional languages corresponding to Afrikaans, Amharic, Catalan, Czech, Danish, Spanish, Indonesian, Icelandic, Italian, Malagasy, Dutch, Polish, Slovak, Swahili, Thai, Turkish, Urdu, Vietnamese, and Yiddish.
\end{itemize}


\begin{table*}[tb!]
\small
\centering
\begin{tabular}{@{}lllllll@{}}
\toprule
src/trgt    & \textbf{de} & \textbf{en} & \textbf{he} & \textbf{ne} & \textbf{hi} & \textbf{ja} \\ \midrule
\textbf{de} & 0.2801      & 0.3177      & 0.2881      & 0.2231      & 0.2685      & 0.3954      \\
\textbf{en} & 0.3401      & 0.2508      & 0.2761      & 0.2238      & 0.2615      & 0.3927      \\
\textbf{he} & 0.3527      & 0.3295      & 0.2612      & 0.2536      & 0.2912      & 0.4041      \\
\textbf{ne} & 0.3255      & 0.2861      & 0.2716      & 0.1510       & 0.2531      & 0.3887      \\
\textbf{hi} & 0.3221      & 0.2981      & 0.2873      & 0.2415      & 0.2083      & 0.4045      \\
\textbf{ja} & 0.373       & 0.3536      & 0.3194      & 0.2825      & 0.3232      & 0.3869      \\ \bottomrule
\end{tabular}
\caption{Averaged MRR scores for five seeds. Bilingual training was done with five seeds over a group of six diverse languages to verify the results are stable. The table shows mean results. The column indicates the source languages, the row indicates the target languages.}
\label{tab:pretrain-mean-seeds}
\end{table*}


\begin{table*}[tb!]
\small
\centering
\begin{tabular}{@{}lllllll@{}}
\toprule
src/trgt    & \textbf{de} & \textbf{en} & \textbf{he} & \textbf{ne} & \textbf{hi} & \textbf{ja} \\ \midrule
\textbf{de} & 0.028       & 0.0031      & 0.0118      & 0.0013      & 0.0103      & 0.0068      \\
\textbf{en} & 0.0062      & 0.0229      & 0.0061      & 0.0054      & 0.0071      & 0.0086      \\
\textbf{he} & 0.0037      & 0.0036      & 0.0113      & 0.0019      & 0.0051      & 0.0023      \\
\textbf{ne} & 0.0287      & 0.0049      & 0.0047      & 0.0041      & 0.0094      & 0.0078      \\
\textbf{hi} & 0.0033      & 0.0035      & 0.0113      & 0.0276      & 0.0531      & 0.0196      \\
\textbf{ja} & 0.0057      & 0.0059      & 0.0062      & 0.0066      & 0.0061      & 0.0029  \\  \bottomrule 
\end{tabular}
\caption{standard deviations for MRR scores over five seeds. Bilingual training was done with five seeds over a group of six diverse languages to verify the results are stable. The table shows the standard deviation of the results. The column indicates the source languages, the row indicates the target languages.}
\label{tab:pretrain-std-seeds}
\end{table*}

\paragraph{Transfer Distribution}
\label{sec:transfer-distribution}

\begin{figure}[tb!]
    \centering
    \includegraphics[width=\linewidth]{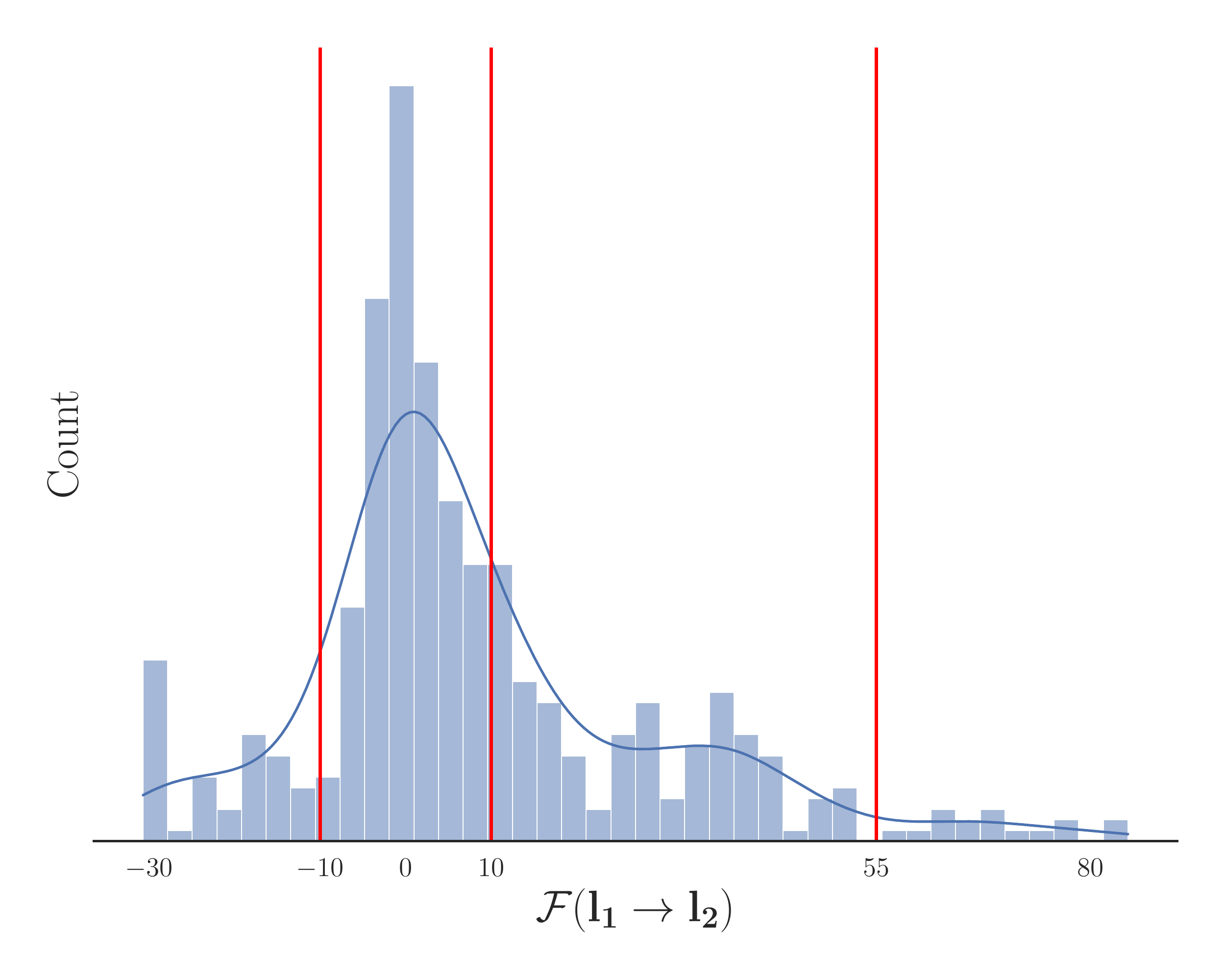}
    \caption{Histogram of cross-lingual transfers $\ftt{l_i}{l_j}$. Horizontal lines (at $-10$, $10$, and $55$) are the borders between four transfer levels.}
    \label{fig:transfer_histogram}
\end{figure}

In the histogram of cross-lingual transfers (Figure~\ref{fig:transfer_histogram}), we observe that the distribution has multiple local maximums (modes).  
We distinguish four main level of cross-lingual transfer described in Section~\ref{subsec:language-graph} ($\ftt{l_i}{l_j}$):
\begin{itemize}
    \setlength{\itemsep}{0.5pt}
    \item negative transfer $\ftt{l_i}{l_j} < -10$
    \item neutral transfer $-10 \leq \ftt{l_i}{l_j} < 10$
    \item positive transfer $10 \leq \ftt{l_i}{l_j} < 55$
    \item very positive transfer $55 \leq \ftt{l_i}{l_j}$
\end{itemize}

The choice of division borders was done in order to separate distinct modes of the distribution and to obtain interpretable bins (e.g. neutral transfer centered around zero).



 \begin{table*}[tb!]
 \tiny
 \centering
 \resizebox{\textwidth}{!}{%
 \begin{tabular}{ccccccccccccccccccccc}
 \hline
 \textbf{my} & \textbf{ne} & \textbf{de}                  & \textbf{hi}                  & \textbf{en}                  & \textbf{hu} & \textbf{hy}                  & \textbf{ar}                  & \textbf{he} & \textbf{ru}                  & \textbf{zh} & \textbf{ta}                  & \textbf{ko} & \textbf{ga} & \textbf{ja} & \textbf{fi}                  & \textbf{cy} & \textbf{te} & \textbf{el}                  & \textbf{fr} & \textbf{pms} \\ \hline
 23.3        & 24.2        & \textcolor{reciepientblue}{24.7} & \textcolor{reciepientblue}{25.0} & \textcolor{reciepientblue}{25.9} & 27.2        & \textcolor{reciepientblue}{28.3} & \color{donorred}{32.1} & 33.4        & \color{donorred}{34.3} & 36.3        & \color{donorred}{36.4} & 36.5        & 36.7        & 37.6        & \color{donorred}{38.0} & 39.0        & 40.0        & \color{donorred}{41.0} & 41.4        & 58.9         \\ \hline
 \end{tabular}%
 }
 \caption{Monolingual results (MRR scores) for all 22 languages in our study, ordered from low to high. Colors coding follows Figure~\ref{fig:lang-type} where \emph{O type languages} are marked in \textcolor{donorred}{red} and \emph{AB+ type languages} languages are marked in \textcolor{reciepientblue}{blue}. Monolingual performance explains some of the pretraining contribution, namely recipient languages appear near the low end of the spectrum while donors appear towards the end.}
 \label{tab:mono-res}
 \end{table*}

\end{document}